# A Dialogue-Based Human-Robot Interaction Protocol for Wheelchair and Robotic Arm Integrated Control

Guangping Liu, Nicholas Hawkins, Billy Madden, Tipu Sultan, Madi Babaiasl

Saint Louis University

## INTRODUCTION

Despite extensive research on control interfaces for wheelchairs and wheelchair-mounted robotic arms (WMRAs), such as joysticks [1], touchscreens [2], and predefined commands [3], operating a coupled wheelchair–arm system often requires frequent mode switching between mobility and manipulation, making interaction unintuitive and cognitively demanding [2,4]. A primary limitation is the "control dimensionality gap": conventional interfaces like 2D joysticks are insufficient for expressing high-dimensional human intent. To overcome these barriers, we hypothesize that natural dialogue or conversations between human beings provide a more intuitive and efficient medium for controlling assistive robots by allowing for high-dimensional semantic expression. To evaluate this hypothesis, we evaluate people's feedback on using a human-robot conversational interaction comparing with traditional manual control, where we use joysticks for the wheelchair and game controllers for the robotic arm. In this pilot study, we propose a Wizard-of-Oz (WoZ) interaction protocol that simulates dialogue control of a wheelchair and robotic arm with five activities of daily-living (ADLs) including cleaning, drinking, feeding, drawer opening, and door opening. The results indicate that most participants preferred the dialogue-based interaction and found it easy to use.

## METHODS

### WheelArm Hardware Setup

We developed our testbed of a WMRA and a wheelchair with Kinova Gen3 6-DoF robotic arm and WHILL Model CR2 wheelchair, named WheelArm (**Figure 1**). Two laptops, the Dell Precision 5570 (Core i9, 32 GB RAM) and the Dell Precision 7780 (Core i9, 64 GB RAM) receive data from Meta Quest 3S (128 GB) for teleoperation. An ego-centered camera (Luxions OAK-D W) and a wrist camera (RealSense D415) are set up for perception.

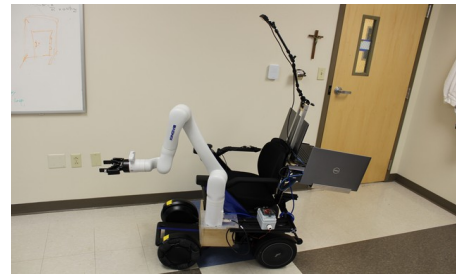

**Figure 1: WheelArm Hardware Overview**

### Assistive Tasks and Environment Setup

The study takes place in Rooms A and B. Room A is the experiment area and is arranged with a custom-built door, a drawer, a feeding/drinking table, and a cleaning setup consisting of a table with trash and a trash bin. Room B functions as the teleoperation area and includes a workstation with a microphone (**Figure 2**).

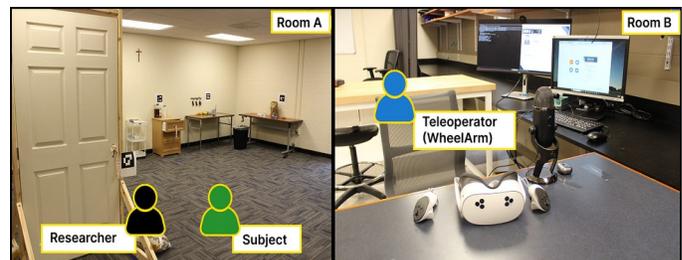

**Figure 2: Environment Setup**

### Teleoperation Framework

Our teleoperation framework is developed on OpenTeach [5], an open-source teleoperation framework for manipulation. We integrated a robotic arm with a wheelchair platform, extending manipulation to include navigation, to build the resulting WheelArm teleoperation system. Compared with the original OpenTeach, we streamlined the interface by replacing hand tracking with controller tracking: the left controller poses and its buttons drive the robotic arm motion and gripper operations, while the right controller's joystick commands the wheelchair. To enable the WoZ protocol, we further augmented the VR application with a researcher-oriented Graphic User Interface (GUI), allowing the teleoperator to start and stop data collection independently without coordinating with the on-site researcher.

### Wizard-of-Oz Protocol

To capture natural user responses, we use WoZ, a study method in which participants engage with a system that appears autonomous, while a human operator discreetly controls its behavior behind the scenes. In our study, we simulate WheelArm's dialogue-driven autonomy to enable participants to experience realistic interaction without requiring a fully autonomous assistive robot.

We implement WoZ using a two-researcher setup (**Figure 2**). During each session, one researcher acts as the teleoperator in Room B, while a second researcher stays with the participant in Room A. This physical separation keeps the teleoperator out of sight, and participants are informed that WheelArm is an intelligent robot capable of spoken dialogue and our experiments are for data collection. All communication is carried over a shared Zoom audio channel, and the teleoperator's voice is transformed in real time using the w-okada system [6] to mask speaker-specific cues and preserve a consistent robot persona.

**Table 1. Researcher-System Handshake Dialogue**

| Researcher | Hi WheelArm, launch the WheelArm Intelligent System. |
|---|---|
| Teleoperator (WheelArm) | System launched. |
| Researcher | Hi WheelArm, name the data collection file please. |
| Teleoperator (WheelArm) | File name sent. |
| Researcher | Hi WheelArm, please start data collection. |
| Teleoperator (WheelArm) | Data collection started. Hi I'm WheelArm, how can I help you today? |

Communication between the on-site researcher and the teleoperator follows the protocol in **Table 1**. This protocol structures their coordination while maintaining the illusion that WheelArm is autonomous, so participants believe they are interacting with an intelligent robot.

Experiment Procedures

The teleoperator was trained by senior researchers over 30 hours in preparation to perform the tasks reliably and consistently. This preparation improves motion quality and minimizes teleoperation errors. To better reflect robot-like partial observability, the teleoperator can only see the scene through the wrist-mounted and egocentric camera feeds in the VR interface, restricting awareness to onboard sensing and encouraging robot-plausible actions and clarifying questions.

Our pilot study was conducted with five participants with no disabilities (N=5) to evaluate the fundamental usability and safety of the dialogue-driven control framework before clinical implementation with target end-users. This experiment follows an approved Institutional Review Board (IRB). Following IRB-approved screening, eligible participants schedule an in-person visit to Room A. At the start of the session, they review and sign the consent form, HIPAA authorization, and a Demographic and Background Questionnaire. The researcher then introduces WheelArm, presented as an intelligent, autonomous assistive robot, and briefs participants on the five study tasks. After the data collection phase, participants are debriefed and informed of the WoZ procedure, then asked to re-consent with the option to withdraw their data. Those who re-consent, complete a short manual-control door-opening trial using a manual control with a Xbox controller and a joystick, and finally fill out the post-study feedback questionnaire.

The questionnaire is organized into two Likert-scale table and an open-ended question session in **Table 2**. A five-item Likert-scale section measures participants' enjoyment of the dialogue-driven interaction while the other one probes perceived autonomy and asks participants to compare WheelArm's behavior against a manual-control baseline (joystick and Xbox controller). The open-ended questions collect qualitative feedback, including comments and suggestions about the interaction and autonomy.

**Table 2: Questionnaire Likert-scale Table**

| | |
|---|---|
| Section 1: Dialogue-based HRI with the WheelArm | I enjoy interacting with WheelArm through dialogue. |
| | The conversation-based interaction allowed me to control WheelArm intuitively and effectively. |
| | I am satisfied with the level of intelligent autonomy demonstrated by WheelArm. |
| | I trust WheelArm when I'm using it. |
| | How likely are you to use this interaction method when operating WheelArm? |
| Section 2: WheelArm Execution Human Preference | Compared with manual control (joystick/Xbox), autonomous WheelArm control is easier to use. |
| | Compared with manual control, autonomous WheelArm control is more effective at completing tasks. |
| | Given the choice, I would prefer to use autonomous control rather than manual control for the WheelArm. |

| Section 3: Open-ended Questions | Do you have any suggestions for this WheelArm autonomy? |
|---|---|
| | What do you like or dislike about this conversational interaction? |
| | What other tasks do you like WheelArm to be able to perform? |

Data Processing

In questionnaire analysis, mapping to scores of 5 through 1, all Likert-scale table items use a 5-point scale spanning strongly agree/very likely to strongly disagree/very unlikely. For each question, we summarize responses using the median. We also report the top-box rate, the share of ratings in the two highest categories (4–5), to capture the proportion of strongly positive responses. To evaluate the internal consistency of the Likert-scale tables, we compute Cronbach's alpha (α), a standard reliability metric that measures how well the items in a scale cohere as a set. Values above 0.7 are typically considered well-correlated, whereas values below 0.5 suggest poor consistency and that the items may not be measuring the same thing. Qualitative responses from Section 3 (**Table 2**), which is an open-ended question section, are analyzed using an inductive thematic approach to identify topics in user preferences and suggestions.

## RESULTS AND DISCUSSION

Subjects Information

### Table 3. Subjects Demographic Overview

| Total Participants(N) | 5 |
|---|---|
| Age (Mean±SD) | 22.8±3.92 |
| Gender | 2F/3M |
| Prior Assistive Technology Experience | None |

In the study, we recruited 5 subjects, 3 males and 2 females. Subjects' demographic information is listed in the **Table 3**.

Liker-scale Table Cronbach's Alpha Analysis

The Likert-scale analysis indicates that our evaluation items function as intended and that participants responded positively to the dialogue-driven interaction and WheelArm's perceived autonomy. Using Cronbach's alpha (α) to assess internal consistency, the first Likert table measuring participants' acceptance of the dialogue-based interaction, achieved a high reliability (α = 0.87). In contrast, the second Likert table, which assesses participants' preferences between WheelArm's automated behavior and manual control using a joystick and an Xbox controller, showed low reliability (α = 0.26). This suggests that responses to the first table were highly consistent, whereas the items in the second table were not perceived as measuring a single underlying construct. Several factors may contribute to this discrepancy. Additionally, inspecting **Figure 3** supports this interpretation: while most participants rated WheelArm's autonomy as easier than manual control, more responses clustered around undecided for autonomy efficiency and overall preference. One plausible explanation is that participants have no disabilities and may have expected that, with practice, they could perform manual control effectively, making them less willing to strongly prefer autonomy despite acknowledging its usability advantages.

Likert-scale Table Score Analysis

To explore subjects' feedback on the dialogue-based interaction and WheelArm autonomy, **Figure 3** shows that the median ratings for questions including effective control, trust autonomy, likely to use in daily life, autonomy easier than manual, autonomy more effective, and prefer autonomy overall are at agree/likely. Two questions, enjoyed interaction and satisfied with autonomy are of strong agree/very likely. This indicates consistently positive attitudes across participants. The remaining questions also trend toward agree or likely, suggesting that participants generally found the interaction understandable and acceptable during the study.

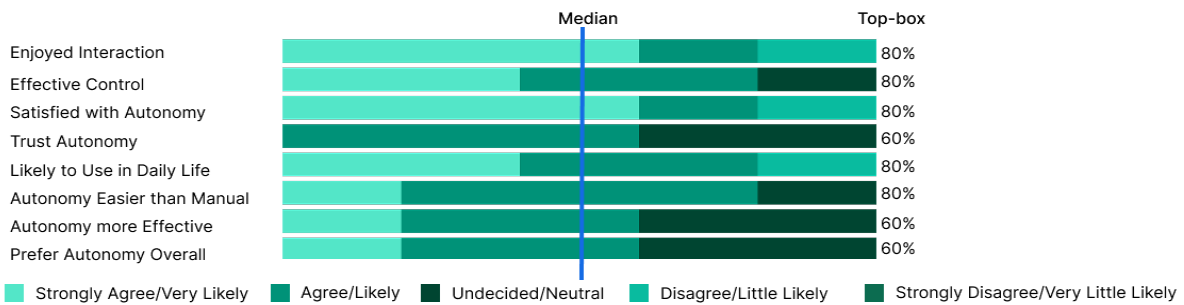

Figure 3: Likert-scale Table Response Distribution

The top-box analysis reinforces this pattern. Five items reached an 80% top-box rate (ratings of 4–5), and three additional items reached 60%, meaning that a majority of participants selected one of the two most positive options for most questions. Taken together, the median and top-box results indicate not only overall approval but also a strong concentration of high ratings despite the small sample size.

Open-ended Questions Thematic Analysis

The thematic analysis of the open-ended questions was consistent with the Likert-scale tables (**Figure 4**). Enjoyment of the dialogue-based interface was mentioned four times, often in reference to WheelArm's ability to understand different words and expressions. Participants also proposed expanding the range of assistive tasks: suggestions such as cooking, washing dishes, and opening drinks were raised four times, reflecting interest in broader daily-living support. At the same time, one participant noted that the robot should respond faster, highlighting speed as a practical factor that may influence perceived usability and satisfaction in future iterations.

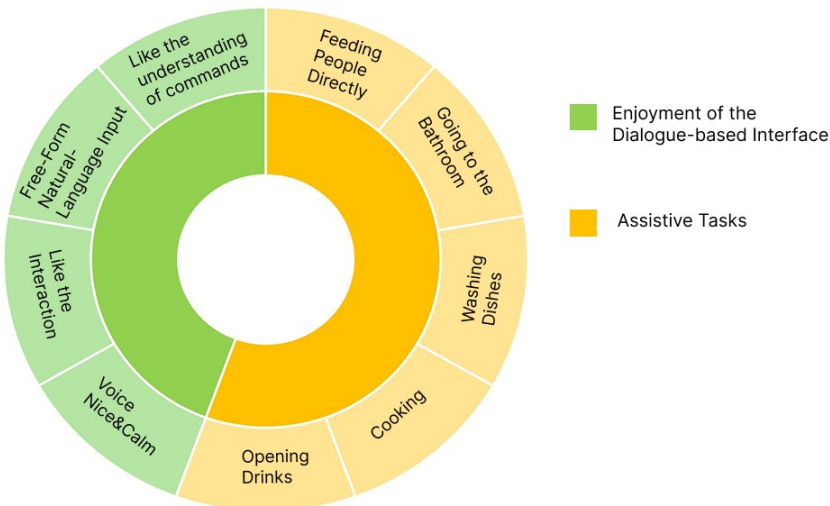

Figure 4: Thematic Analysis

## CONCLUSION

This study demonstrates that natural dialogue effectively bridges the "dimensionality gap" in assistive robotics, providing a more intuitive interface for controlling complex wheelchair-arm systems compared to traditional manual controls. Future work will focus on developing an integrated WheelArm control with the dialogue-based interaction protocol.